\documentclass[journal]{IEEEtran}
\usepackage{epsfig}
\usepackage{graphicx}
\usepackage{amsmath}
\usepackage{amssymb}
\usepackage{amsfonts}
\usepackage{CJK}
\usepackage{amsmath}
\usepackage{algorithm}
\usepackage{algorithmic}
\usepackage{bm}
\usepackage{indentfirst}
\usepackage{multirow}
\usepackage{graphicx}
\usepackage{array}
\usepackage{mathrsfs}
\usepackage{color}
\usepackage{url}

\ifCLASSINFOpdf
\else
\fi
\begin{document}
%
\title{Dual Asymmetric Deep Hashing Learning}
%
%
%

\author{Jinxing Li,
        Bob Zhang,~\IEEEmembership{Member,~IEEE,}
        Guangming Lu,
        David Zhang*,~\IEEEmembership{Fellow,~IEEE}
\thanks{J. Li is with the Department
of Computing, Hong Kong Polytechnic University, Hung Hom, Kowloon (e-mail: csjxli@comp.polyu.edu.hk).}
\thanks{B. Zhang is with the Department of Computer and Information Science, University of Macau, Avenida da Universidade, Taipa, Macau (e-mail: bobzhang@umac.mo).}
\thanks{G. Lu is with Department of Computer Science, Harbin Institute of Technology Shenzhen Graduate School, Shenzhen, China (e-mail: luguangm@hit.edu.cn).}
\thanks{D. Zhang is with the Department
of Computing, Hong Kong Polytechnic University, Hung Hom, Kowloon (e-mail: csdzhang@comp.polyu.edu.hk).}
}

%
%

\markboth{Journal of IEEE Transactions on XX,~Vol.~X, No.~X, XX~2018}%
{Shell \MakeLowercase{\textit{et al.}}: Bare Demo of IEEEtran.cls for IEEE Journals}
%



\maketitle

\begin{abstract}
Due to the impressive learning power, deep learning has achieved a remarkable performance in supervised hash function learning. In this paper, we propose a novel asymmetric supervised deep hashing method to preserve the semantic structure among different categories and generate the binary codes simultaneously. Specifically, two asymmetric deep networks are constructed to reveal the similarity between each pair of images according to their semantic labels. The deep hash functions are then learned through two networks by minimizing the gap between the learned features and discrete codes. Furthermore, since the binary codes in the Hamming space also should keep the semantic affinity existing in the original space, another asymmetric pairwise loss is introduced to capture the similarity between the binary codes and real-value features. This asymmetric loss not only improves the retrieval performance, but also contributes to a quick convergence at the training phase. By taking advantage of the two-stream deep structures and two types of asymmetric pairwise functions, an alternating algorithm is designed to optimize the deep features and high-quality binary codes efficiently. Experimental results on three real-world datasets substantiate the effectiveness and superiority of our approach as compared with state-of-the-art.
\end{abstract}

\begin{IEEEkeywords}
deep learning, hashing learning, image retrieval, similarity
\end{IEEEkeywords}

%
\IEEEpeerreviewmaketitle

\section{Introduction}
%
%
%
%
\IEEEPARstart{W}{ith} the rapid growth of multimedia data in search engines and social networks, how to store these data and make a fast search when an novel one such as the image is given, plays a fundamental role in machine learning. Due to the low storage cost and fast retrieval speed, hashing techniques have attracted much attention and are widely applied in nearest neighbor search \cite{andoni2006near} for information retrieval on large scale datasets. Hashing learning aims to project the data from the original space into a Hamming space by generating compact codes. These codes can not only dramatically reduce the storage overhead and achieve a constant or sub-linear time complexity in information search, but also preserve the semantic affinity existing in the original space.

Many hashing methods have been studied \cite{gionis1999similarity} \cite{kulis2009kernelized} \cite{gong2013iterative} \cite{lin2014fast} \cite{weiss2009spectral} \cite{liu2011hashing} \cite{liu2014discrete} \cite{liu2012supervised} \cite{shen2015supervised}. Generally, these approaches can be roughly classified into two categories: data-independent and data-dependent hashing methods. Locality Sensitive Hashing (LSH) \cite{gionis1999similarity} and its extension Kernelized LSH (KLSH) \cite{kulis2009kernelized}, as the most typical data-independent hashing methods, were proposed to obtain the hashing function by using random projections. Although the designation of these data-independent methods is quite simple, they often meet a performance degradation when the length of the binary codes is relatively low. By contrary, instead of randomly generating the hashing function like LSH does, data-dependent methods aims to learn a data-specific hashing function by using the training data, being capable of generating shorter binary codes but achieving more remarkable performance. Therefore, various data-dependent hashing approaches containing both unsupervised and supervised have been proposed. Unsupervised hashing, e.g. Spectral Hashing \cite{weiss2009spectral}, Anchor Graph Hashing (AGH) \cite{liu2011hashing}, and Discrete Graph Hashing (DGH) \cite{liu2014discrete} etc., only try to utilize the data structure to learn compact binary codes to improve the performance. By taking the label information into account, supervised hashing methods attempt to map the original data into a compact Hamming space to preserve the similarity between each pair samples. Many representative works including Fast Supervised Hashing (FastH) \cite{lin2014fast}, Kernel Supervised Hashing (KSH) \cite{liu2012supervised}, and Supervised Discrete Hashing (SDH) \cite{shen2015supervised} etc., demonstrate that supervised hashing methods often obtain an outstanding performance compared with unsupervised hashing methods. Thus, we focus on studying the supervised hashing method in this paper.

Although some traditional supervised hashing methods achieve a good performance in some applications, most of them only linearly map the original data into a Hamming space by using the hand-crafted features, limited their application for large-scale datasets which have complex distributions. Fortunately, due to the powerful capability of data representation, deep learning \cite{krizhevsky2012imagenet} \cite{simonyan2014very} provides a promising way to jointly represent the data and learn hash codes. Some existing deep learning based hashing methods have been studied, such as Deep Supervised Hashing (DSH) \cite{liu2016deep} and Deep Pairwise Supervised Hashing (DPSH) \cite{li2015feature}, etc. These approaches demonstrate the effectiveness of the end-to-end deep learning architecture for hashing learning.

Despite the wide applications of deep neural network on hashing learning, most of them are symmetric structures in which the similarity between each pair points are estimated by the Hamming distance between the outputs of the same hash function \cite{shen2017deep}. As described in \cite{shen2017deep}, a crucial problem is that this symmetric scheme would result in the difficulty of optimizing the discrete constraint. Thus in this paper, we propose a novel asymmetric hashing method to address aforementioned problem. Note that a similar work was described by Shen et al. \cite{shen2017deep}, named deep asymmetric pairwise hashing (DAPH). However, our study is quite distinctive from DAPH. Shen et al. tried to approximate the similarity affinity by exploiting two different hashing functions, which can preserve more similarity information among the real-value features. However, DAPH only exploits a simple Euclidean distance, but ignores the semantic structure between the learned real-value features and binary codes \cite{da2017amvh} \cite{jiang2017asymmetric}. One major deficiency is that it is difficult to efficiently preserve the similarity in the learned hash functions and discrete codes. Furthermore, in DAPH, two different types of discrete hash codes corresponding to two hash functions are estimated in the training time. However this strategy would enlarge the gap between two schemes, resulting in a performance degradation. By contrast, we not only propose a novel asymmetric structure to learn two different hash functions and one consistent binary code for each sample at the training phase, but also asymmetrically exploit real-value and multiple integer values, which permits the better preservation of similarity between the learned features and hash codes. Experiments show that this novel asymmetric structure can get a better performance in image retrieval and quicker convergence at the training stage.

The main contributions of the proposed method are shown as follows:\\
(1) A novel asymmetric deep structure are proposed. Two streams of deep neural networks are trained to asymmetrically learn two different hash functions. The similarity between each pair images are utilized through a pairwise loss according to their semantic/label information.\\
(2) The similarity between the learned features and binary codes are also revealed through an additional asymmetric loss. Real-value features and binary codes are bridged through an inner product, which alleviates the binary limitation, better preserves the similarity, and speeds up convergence at the training phase.\\
(3) By taking advantage of these two asymmetric properties, an alternative algorithm is designed to efficiently optimize the real values and discrete values.\\
(4) Experimental results on three large-scale datasets substantiate the effectiveness and superiority of our approach as compared with some existing state-of-the-art hashing methods in
image retrieval.

The rest of this paper is organized as follows. In Section 2, the related works including data-independent and data-dependent hashing methods are briefly reviewed. In Section 3, the proposed \textbf{D}ual \textbf{A}symmetric \textbf{D}eep \textbf{H}ashing Learning (DADH) is then analyzed, followed by its optimization. In Section 4, experiments are conducted on three real-world datasets, and some comparisons, parameter sensitivity analysis and convergence analysis are discussed. This paper is finally concluded in Section 5.

\section{Related Works}
As mentioned before, the hashing method can be roughly separated into data-independent and data-dependent hashing.

Locality Sensitive Hashing (LSH) \cite{gionis1999similarity} aims to use several hash functions to randomly project the data into a Hamming space, so as to ensure the probability of collision is much higher for data points which are close to each other than for those which are far apart. Consider the non-linearity existing in many real-world datasets, LSH was generated to accommodate arbitrary kernel functions (KLSH) in \cite{kulis2009kernelized}. Some other priors, such as $p$-stable distributions \cite{datar2004locality} and shift-invariant kernels \cite{raginsky2009locality}, are also embedded to extend LSH for performance improvement.

Different from data-independent methods, the data-dependent methods try to learn more compact codes from a given dataset to achieve a satisfactory search accuracy. According to whether the label information is available, data-dependent hashing can also be classified into unsupervised and supervised. Typical learning criteria for unsupervised hashing methods contains graph learning \cite{weiss2009spectral} \cite{liu2011hashing} \cite{liu2014discrete} \cite{jiang2015scalable} and error minimization \cite{gong2013iterative} \cite{jegou2011product} \cite{shen2013inductive}. \emph{Graph Learning:} Yair et al. \cite{weiss2009spectral} proved that finding a best code is associated with the problem of graph partitioning. Thus, a spectral hashing (SH) was proposed to learn the hash function. Another graph hashing named Anchor Graph Hashing (AGH) was presented by Liu et al. \cite{liu2011hashing}, which is capable of capturing the neighborhood structure inherent automatically. In order to avoid the high complexity of existing graph hashing methods, Jiang et al. \cite{jiang2015scalable} proposed a scalable graph hashing (SGH) which can be effectively applied to the large-scale dataset search. Although SH, AGH and SGH achieve a satisfactory performance in some datasets, both of them relax the optimization by discarding the discrete constraints, which results in an accumulated quantization error. To address this problem, a discrete graph hashing (DGH) \cite{liu2014discrete} was proposed, which can find the neighborhood structure inherent in a discrete code space. \emph{Error Minimization:} A typical method is the iterative quantization (ITQ) \cite{gong2013iterative} which aims to project the data to the vertices of a binary hypercube and minimize the quantization error. Additionally, the method in \cite{jegou2011product} makes a quantization by decomposing the input space into a Cartesian product of low-dimensional subspaces, dramatically reducing the quantization noise. Different from most single-bit quantization, a double-bit quantization hashing \cite{kong2012double} was also studied by quantizing each dimension into double bits.

In contrast to unsupervised hashing methods, supervised hashing learning utilizes the label information to encourage the binary codes in the Hamming space to preserve the semantic relationship existing in the raw data. For instance, Mohammad et al. \cite{norouzi2011minimal} introduced a hinge-like loss function to exploit the semantic information. Besides, Li et al. \cite{zhang2014supervised} projected the raw data into a latent subspace, and the label information is embedded on this subspace to preserve the semantic structure. The Jensen Shannon Divergence is also utilized in \cite{fan2013supervised} to learn the binary codes within a probabilistic framework, in which an upper bound is derived for various hash functions. Being similar to KLSH, Liu et al. \cite{liu2012supervised} proposed a supervised hashing with kernels, in which the similar pairs are minimized while the dissimilar pairs are maximized. Consider the discrete constraint, the supervised discrete hashing (SDH) \cite{shen2015supervised} was proposed to not only preserve the semantic structure, but also discretely learn the hash codes without any relaxation. However, this discrete optimization is time-consuming and unscalable. To tackle this problem, a novel method named column sample based discrete supervised hashing (COSDISH) was presented to directly obtain the binary codes from semantic information.

Although various works mentioned above have been studied, they only project the data into the Hamming space by using the hand-crafted features. The main limitation is that they would meet a performance degradation if the distribution of a real-world dataset is complex. Fortunately, deep learning provides a reasonable and promising solution. Liong et al. \cite{erin2015deep} used the deep structure to hierarchically and non-linearly learn the hash codes. Convolutional neural network (CNN) was first applied by Xia and Yan et al. to the hashing learning (CNNH) \cite{xia2014supervised}, which simultaneously represents the image and learns a hash function. A novel deep structure \cite{lai2015simultaneous} was then proposed by modifying the fully-connected layer in CNNH to a divide-and-encode module, in which the hash codes can be obtained bit by bit. Also, Can et al. \cite{cao2017deep} combined the quantization model with deep structure to gain a satisfactory performance in image retrieval. Different from the triple loss used in some deep hashing methods, Li et al. \cite{li2015feature} studied a pairwise loss (DPSH) which can effectively preserve the semantic information between each pair outputs. Due to the power of asymmetric structure, the asymmetric deep hashing was also studied in recent years. For instance, Shen et al. \cite{shen2017deep} (DAPH) tried to learn hash functions in an asymmetric network. However, DAPH only exploits two streams to preserve the pairwise label information between the deep neural network outputs, but ignores the similarity between the real-value features and binary codes. Thus, in this paper, we propose a novel deep hashing method to not only exploit the label information between each two outputs through an asymmetric deep structure, but also semantically associated the learned real-value features with the binary codes.

\begin{figure*}
\begin{center}
   \includegraphics[width=0.85\linewidth]{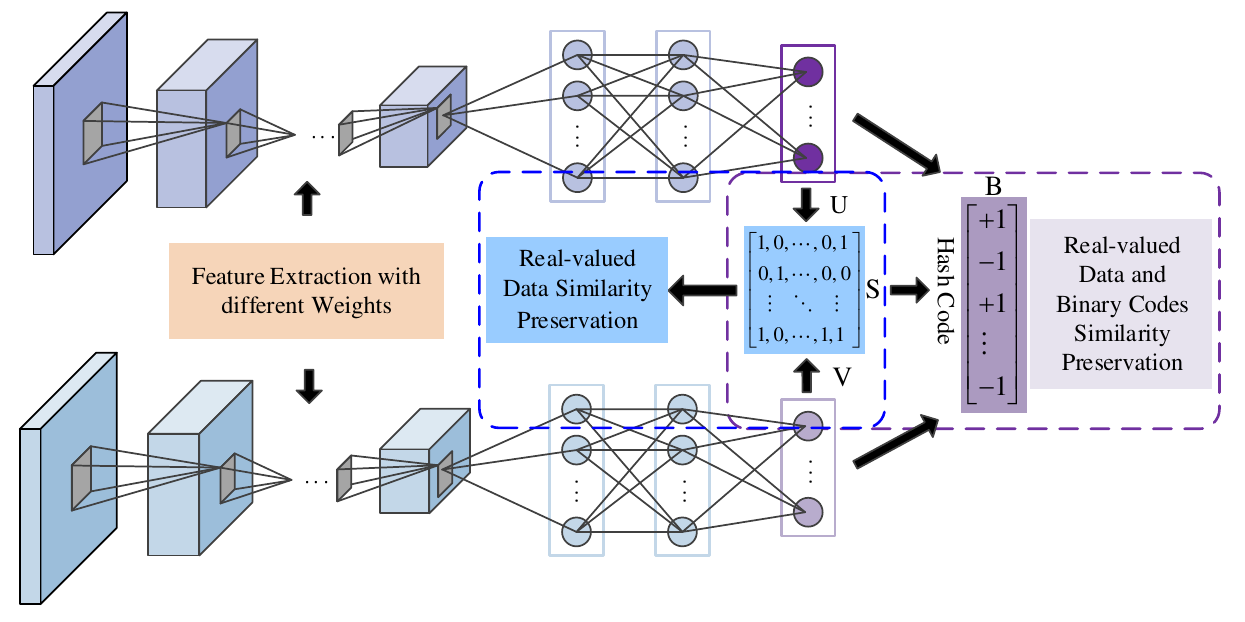}
\end{center}
   \caption{The framework of the proposed method. Two streams with five convolution layers and two full-connected layers are used for feature extraction. For the real-valued outputs from these two neural networks, their similarity is preserved by using a pairwise loss. Based on the outputs, a consistent hash code is generated. Furthermore, an asymmetric loss is introduced to exploit the semantic information between the binary code and real-valued data.}
\label{fig:DADH}
\end{figure*}

\section{The Proposed Method}
In this section, we first give some notations used in this paper, as well as the problem definition. The proposed Dual Asymmetric Deep Hashing Learning (DADH) is then described, followed by its optimization.

\subsection{Notation and Problem Definition}
In this paper, since there are two streams in the proposed method, we use the uppercase letters $\mathbf{X} = \{\mathbf{x}_{1},\cdots , \mathbf{x}_{i},\dots,\mathbf{x}_{N}\} \in \mathbb{R}^{N \times d_{1} \times d_{2} \times 3}$ and $\mathbf{Y} = \{\mathbf{y}_{1},\cdots , \mathbf{y}_{i},\dots,\mathbf{y}_{N}\} \in \mathbb{R}^{N \times d_{1} \times d_{2} \times 3}$ to denote the input images in the first and second deep neural networks, respectively, where $N$ is the number of training samples, $d_{1}$ and $d_{2}$ are the length and width for each image. Note that, although $\mathbf{X}$ and $\mathbf{Y}$ are represented with different symbols, both of them denote the same training data. In our experiments, we only alternatively use training samples $\mathbf{X}$ and $\mathbf{Y}$ in the first and second networks. Since our method is supervised learning, the label information can be used. Let the uppercase letter $\mathbf{S} \in \{-1,+1\}$ denote the similarity between $\mathbf{X}$ and $\mathbf{Y}$ and $S_{i,j}$ is the element in the $i$-th row and $j$-th column in $\mathbf{S}$. Let $S_{i,j}=1$ if $\mathbf{x}_{i}$ and $\mathbf{y}_{j}$ share the same semantic information or label, otherwise $S_{i,j}=-1$.

Denote the binary codes as $\mathbf{B} = [\mathbf{b}_{1},\cdots , \mathbf{b}_{i},\dots,\mathbf{b}_{N}]^{T} \in \mathbb{R}^{N \times k}$ and the $k$-bit binary code of the $i$-th sample as $\mathbf{b}_{i} \in \{-1,+1\}^{k\times 1}$. The purpose of our model is to learn two mapping functions $\mathcal{F}$ and $\mathcal{G}$ to project $\mathbf{X}$ and $\mathbf{Y}$ into the Hamming space $\mathbf{B}$. $\mathbf{b}_{i} = $ sign$(\mathcal{F}(\mathbf{x}_{i}))$ and $\mathbf{b}_{j} = $ sign$(\mathcal{G}(\mathbf{y}_{j}))$, where sign$(\cdot)$ is an element-wise sign function, and sign($x$) = 1 if $x\geq 0$, otherwise sign(x) = -1. The Hamming distance  $\mathbf{dist}_{H}(\mathbf{b}_{i},\mathbf{b}_{j})$ between $\mathbf{b}_{i}$ and $\mathbf{b}_{j}$ should be as small as possible if $s_{ij}=1$ and vice versa. Due to the power of deep neural network in data representation, we apply the convolution neural work to learn the hash functions. Specifically, the CNN-F structure \cite{chatfield2014return} is adopted to perform feature learning. In CNN-F model, there are eight layers including five convolutional layers as well as three fully-connected layers. The network structure is listed in Table \ref{tab:cnnf}, where "f." means the filter, "st." means the convolution stride, "LRN" means the Local Response Normalization \cite{krizhevsky2012imagenet}. In order to get the final binary code, we replace the last layer in CNN-F with a $k$-D vector and the k-bit binary codes are obtained through a \textup{sign} operation on the output of the last layer. In this paper, CNN-F model is applied to both streams in our proposed asymmetric structure.

\begin{table}\small
\caption{The network structure of CNN-F.}
\centering
\begin{tabular}{|c | c|}
\hline
\multicolumn{1}{|c|}{Layer}&{Structure}\\
\hline
{conv1}&{f. 64 $\times$ 11 $\times$ 11; st. 4$\times$4; pad. 0; LRN.; $\times$2 pool}\\
\hline
{conv2}&{f. 265 $\times$ 5 $\times$ 5; st. 1$\times$1; pad. 2; LRN.; $\times$2 pool}  \\
\hline
{conv3}&{f. 265 $\times$ 3 $\times$ 3; st. 1$\times$1; pad. 1}  \\
\hline
{conv4}&{f. 265 $\times$ 3 $\times$ 3; st. 1$\times$1; pad. 1}  \\
\hline
{conv5}&{f. 265 $\times$ 3 $\times$ 3; st. 1$\times$1; pad. 1; $\times$2 pool}  \\
\hline
{full6}&{4096}  \\
\hline
{full7}&{4096}  \\
\hline
{full8}&{$\rightarrow$ k-bit hash code}  \\
\hline
\end{tabular}\label{tab:cnnf}
\end{table}

\subsection{Dual Asymmetric Deep Hashing Learning}
The main framework of the proposed method is shown in Fig.\ref{fig:DADH}. As we can see, there are two end-to-end neural networks to discriminatively represent the inputs. For a pair of outputs $\mathbf{F}$ and $\mathbf{G}$ in these two streams, their semantic information is exploit through a pairwise loss according to their predefined similarity matrix. Since the purpose is to obtain hash functions through the deep networks, the binary code $\mathbf{B}$ is also generated by minimizing its distance between $\mathbf{F}$ and $\mathbf{G}$. Furthermore, in order to preserve the similarity between the learned binary codes and real-value features, and alleviate the binary limitation, another asymmetric pairwise loss is introduced by using the inner product of the hash codes $\mathbf{B}$ and learned features $\mathbf{F}$ ($\mathbf{G}$).

Denote $\mathbf{f}(\mathbf{x}_{i},\mathbf{W}_{f})\in \mathbb{R}^{k\times 1}$ as the output of the $i$-th sample in the last layer of the first stream, where $\mathbf{W}_{f}$ is the parameter of the network. To simplify the notation, we use $\mathbf{f}_{i}$ to replace $\mathbf{f}(\mathbf{x}_{i},\mathbf{W}_{f})$. Similarly, we can obtain the output $\mathbf{g}_{j}$ corresponding to the $j$-th sample under the parameter $\mathbf{W}_{g}$ in the second network. Thus, the features $\mathbf{F} = [\mathbf{f}_{1},\cdots,\mathbf{f}_{i},\mathbf{f}_{n}]^{T} \in \mathbb{R}^{n\times k}$ and $\mathbf{G} = [\mathbf{g}_{1},\cdots,\mathbf{g}_{i},\mathbf{g}_{n}]^{T} \in \mathbb{R}^{n\times k}$ corresponding to the first and second networks are then gained.

To learn an accurate binary code, we set $\textup{sign}(\mathbf{f}_{i})$ and $\textup{sign}(\mathbf{g}_{i})$ to be close to their corresponding hash code $\mathbf{b}_{i}$. A general way is to minimize the $L_{2}$ loss between them.
\begin{equation}\small
\begin{aligned}
\min &\left \| \textup{sign}(\mathbf{f}_{i}) - \mathbf{b}_{i}\right \|_{2}^{2}+\left \| \textup{sign}(\mathbf{g}_{i}) - \mathbf{b}_{i}\right \|_{2}^{2},\\ &\textup{s.t.} \ \mathbf{b}_{i} \in \{-1,+1\}
\end{aligned}\label{eq:L2}
\end{equation}

However, it is difficult to make a back-propagation for the gradient with respect to $\mathbf{f}_{i}$ or $\mathbf{g}_{i}$ in Eq.(\ref{eq:L2}) since their gradients are zero anywhere. In this paper, we apply $\textup{tanh}(\cdot)$ to softly approximate the $\textup{sign}(\cdot)$ function. Thus, Eq.(\ref{eq:L2}) is transformed into
\begin{equation}
\begin{aligned}
\min &\left \| \textup{tanh}(\mathbf{f}_{i}) - \mathbf{b}_{i}\right \|_{2}^{2}+\left \| \textup{tanh}(\mathbf{g}_{i}) - \mathbf{b}_{i}\right \|_{2}^{2},\\ &\textup{s.t.} \ \mathbf{b}_{i} \in \{-1,+1\}
\end{aligned}\label{eq:tanhL2}
\end{equation}

Furthermore, to exploit the label information and keep a consistent similarity between two outputs $\mathbf{F}$ and $\mathbf{G}$, the negative log likelihood of the dual-stream similarities with the likelihood function is exploited.
\begin{equation}
\begin{aligned}
p(S_{ij}|\textup{tanh}(\mathbf{f}_{i}),\textup{tanh}(\mathbf{g}_{j}))=\left\{\begin{matrix}
\sigma (\Theta_{ij}) & S_{ij}=1\\
 1-\sigma (\Theta_{ij})& S_{ij}=0
\end{matrix}\right.
\end{aligned}\label{eq:likelihood}
\end{equation}
where $\Theta_{ij} = \frac{1}{2}\textup{tanh}(\mathbf{f}_{i}^{T})\textup{tanh}(\mathbf{g}_{j})$, and $\sigma(\Theta_{ij})=\frac{1}{1+e^{-\Theta_{ij}}}$. Therefore, the pairwise loss for these two different outputs is shown as follows.
\begin{equation}
\begin{aligned}
\min -\sum_{i,j=1}^{n}(S_{ij}\Theta_{ij} -\log (1+e^{\Theta_{ij}})
\end{aligned}\label{eq:pairwise}
\end{equation}

Although Eq.(\ref{eq:tanhL2}) achieves to approximate discrete codes and Eq.(\ref{eq:pairwise}) exploits the intra- and inter-class information, the similarity between the binary codes and real-value features is ignored. To tackle this problem, another asymmetric pairwise loss is introduced.
\begin{equation}
\begin{aligned}
\min &\left \| \textup{tanh}(\mathbf{f}_{i}^{T})\mathbf{b}_{j} - kS_{ij} \right \|_{2}^{2}+\left \| \textup{tanh}(\mathbf{g}_{i}^{T})\mathbf{b}_{j} - kS_{ij} \right \|_{2}^{2},\\ &\textup{s.t.} \ \mathbf{b}_{i} \in \{-1,+1\}
\end{aligned}\label{eq:asympairwise}
\end{equation}
In Eq.(\ref{eq:asympairwise}), the similarity between the real-valued data and binary codes is measured by their inner product. It is easy to observe that Eq.(\ref{eq:asympairwise}) not only encourages the $\textup{tanh}(\mathbf{f}_{i})$ ($\textup{tanh}(\mathbf{g}_{i})$) and $\mathbf{b}_{i}$ to be consistent, but also preserve the similarity between them. Additionally, our experiments in Section 4 also prove that this kind of asymmetric inner product can quickly make the network converge to a stable value for the real-valued features and hash codes.

Jointly taking Eq.(\ref{eq:tanhL2}), Eq.(\ref{eq:pairwise}) and Eq.(\ref{eq:asympairwise}) into account, the objective function can be obtained as follows:
\begin{equation}
\begin{aligned}
\min_{
\mathbf{F},\mathbf{G},\mathbf{B}
} L=&\left \| \textup{tanh}(\mathbf{F})\mathbf{B}^{T} - k\mathbf{S} \right \|_{F}^{2}+\left \| \textup{tanh}(\mathbf{G})\mathbf{B}^{T} - k\mathbf{S} \right \|_{F}^{2}\\
&-\tau \sum_{i,j=1}^{n}(S_{ij}\Theta_{ij} -\log (1+e^{\Theta_{ij}})+\\
&+\gamma(\left \| \textup{tanh}(\mathbf{F}) - \mathbf{B}\right \|_{F}^{2}+\left \| \textup{tanh}(\mathbf{G}) - \mathbf{B}\right \|_{F}^{2})\\
&+\eta(\left \| \textup{tanh}(\mathbf{F})^{T}\mathbf{1} \right \|_{F}^{2}+\left \| \textup{tanh}(\mathbf{G})^{T}\mathbf{1} \right \|_{F}^{2})\\ &\textup{s.t.} \ \mathbf{b}_{i} \in \{-1,+1\}
\end{aligned}\label{eq:obj}
\end{equation}
where $\tau$, $\gamma$ and $\eta$ are the non-negative parameters to make a trade-off among various terms. Note that the purpose of the forth term $\left \| \textup{tanh}(\mathbf{F})^{T}\mathbf{1} \right \|_{F}^{2}+\left \| \textup{tanh}(\mathbf{G})^{T}\mathbf{1} \right \|_{F}^{2}$ in the objective function Eq.(\ref{eq:obj}) is to maximize the information provided by each bit \cite{jiang2016deep}. In detail, this term makes a balance for each bit, which encourages the number of -1 and +1 to be approximately similar among all training samples.

\subsection{Optimization}
From the objective function Eq.(\ref{eq:obj}), we can see that the real-valued features as well as the weights in two neural networks $(\mathbf{F},\mathbf{W}_{f})$ / $(\mathbf{G},\mathbf{W}_{g})$, and discrete codes $\mathbf{B}$  need to be optimized. Note that this NP-hard problem is highly non-convex, and it is very difficult to directly get the optimal solutions. In this paper, we design an efficient algorithm to optimize them alternatively. Specially, we update one variable by fixing other variables.
\subsubsection{Update $(\mathbf{F},\mathbf{W}_{f})$ with $(\mathbf{G},\mathbf{W}_{g})$ and $\mathbf{B}$ fixed}
By fixing $(\mathbf{G},\mathbf{W}_{g})$ and $\mathbf{B}$, the objective function Eq.(\ref{eq:obj}) can be transformed to
\begin{equation}
\begin{aligned}
\min_{\mathbf{F}} &\left \| \textup{tanh}(\mathbf{F})\mathbf{B}^{T} - k\mathbf{S} \right \|_{F}^{2}
-\tau \sum_{i,j=1}^{n}(S_{ij}\Theta_{ij} -\log (1+e^{\Theta_{ij}})+\\
&+\gamma(\left \| \textup{tanh}(\mathbf{F}) - \mathbf{B}\right \|_{F}^{2}
+\eta\left \| \textup{tanh}(\mathbf{F})^{T}\mathbf{1} \right \|_{F}^{2}
\end{aligned}\label{eq:objF}
\end{equation}
Then the back-propagation is exploited to update $(\mathbf{F},\mathbf{W}_{f})$. Here denote $\mathbf{U}=\textup{tanh}(\mathbf{F})$ and $\mathbf{V}=\textup{tanh}(\mathbf{G})$. The gradient of the objective function with respect to $\mathbf{f}_{i}$ is
\begin{equation}
\begin{aligned}
\frac{\partial L}{\partial \mathbf{f}_{i}}=\{&\sum_{j=1}^{n}[2\mathbf{b}_{j}(\mathbf{b}_{j}^{T}\mathbf{u}_{i}-kS_{ij})+\frac{\tau}{2}
(\sigma (\Theta_{ij})\mathbf{v}_{j}-S_{ij}\mathbf{v}_{j})]\\
&+2\gamma(\mathbf{u}_{i}-\mathbf{b}_{i})+2\eta\mathbf{U}^{T}\textbf{1}\}\odot (1-\mathbf{u}_{i}^{2})
\end{aligned}\label{eq:updateF}
\end{equation}
where $\odot$ denotes the dot product. After getting the gradient $\frac{\partial L}{\partial \mathbf{f}_{i}}$, the chain rule is used to obtain $\frac{\partial L}{\partial \mathbf{W}_{f}}$, and $\mathbf{W}_{f}$ is updated by using back-propagation.
\subsubsection{Update $(\mathbf{G},\mathbf{W}_{g})$ with $(\mathbf{F},\mathbf{W}_{f})$ and $\mathbf{B}$ fixed}
Similarly, by fixing $(\mathbf{F},\mathbf{W}_{f})$ and $\mathbf{B}$, the back-propagation is exploited to update $(\mathbf{G},\mathbf{W}_{g})$. The gradient of the objective function with respect to $\mathbf{g}_{i}$ is
\begin{equation}
\begin{aligned}
\frac{\partial L}{\partial \mathbf{g}_{i}}=\{&\sum_{j=1}^{n}[2\mathbf{b}_{j}(\mathbf{b}_{j}^{T}\mathbf{v}_{i}-kS_{ij})+\frac{\tau}{2}
(\sigma (\Theta_{ij})\mathbf{u}_{j}-S_{ij}\mathbf{u}_{j})]\\
&+2\gamma(\mathbf{v}_{i}-\mathbf{b}_{i})+2\eta\mathbf{V}^{T}\textbf{1}\}\odot (1-\mathbf{v}_{i}^{2})
\end{aligned}\label{eq:updateG}
\end{equation}
After getting the gradient $\frac{\partial L}{\partial \mathbf{g}_{i}}$, the chain rule is used to obtain $\frac{\partial L}{\partial \mathbf{W}_{g}}$, and $\mathbf{W}_{g}$ is updated by using back-propagation.
\subsubsection{Update $\mathbf{B}$ with $(\mathbf{F},\mathbf{W}_{f})$ and $(\mathbf{G},\mathbf{W}_{g})$ fixed}
By fixing $(\mathbf{F},\mathbf{W}_{f})$ and $(\mathbf{G},\mathbf{W}_{g})$, we can get the following formulation.
\begin{equation}
\begin{aligned}
\min_{
\mathbf{B}
} L(\mathbf{B})&=\left \| \mathbf{U}\mathbf{B}^{T} - k\mathbf{S} \right \|_{F}^{2}+\left \| \mathbf{V}\mathbf{B}^{T} - k\mathbf{S} \right \|_{F}^{2}\\
&+\gamma(\left \| \mathbf{U} - \mathbf{B}\right \|_{F}^{2}+\left \| \mathbf{V} - \mathbf{B}\right \|_{F}^{2}) \ \textup{s.t.} \ \mathbf{b}_{i} \in \{-1,+1\}
\end{aligned}\label{eq:objB}
\end{equation}
Then Eq.(\ref{eq:objB}) can be rewrote as:
\begin{equation}
\begin{aligned}
\min_{
\mathbf{B}
}& L(\mathbf{B})=-2\textup{Tr}[\mathbf{B}(k(\mathbf{U}^{T}\mathbf{S}+\mathbf{V}^{T}\mathbf{S})
+\gamma(\mathbf{U}^{T}+\mathbf{V}^{T}))]\\
 &+\left \| \mathbf{B}\mathbf{U}^{T} \right \|_{F}^{2}+\left \| \mathbf{B}\mathbf{V}^{T} \right \|_{F}^{2}+\textup{const} \ \textup{s.t.} \ \mathbf{b}_{i} \in \{-1,+1\}
\end{aligned}\label{eq:objBR}
\end{equation}
where '$\textup{const}$' means a constant value without any association with $\mathbf{B}$. For the sake of simplicity, let $\mathbf{Q} =-2k(\mathbf{S}^{T}\mathbf{U}+\mathbf{S}^{T}\mathbf{V})
-2\gamma(\mathbf{U}+\mathbf{V})$. Eq.(\ref{eq:objBR}) can be simplified to
\begin{equation}
\begin{aligned}
\min_{
\mathbf{B}
}L(\mathbf{B})=\left \| \mathbf{B}\mathbf{U}^{T} \right \|_{F}^{2}+\left \| \mathbf{B}\mathbf{V}^{T} \right \|_{F}^{2}+\textup{Tr}[\mathbf{B}\mathbf{Q}^{T}] +\textup{const}\\ \ \textup{s.t.} \ \mathbf{b}_{i} \in \{-1,+1\}
\end{aligned}\label{eq:objBR2}
\end{equation}

According to Eq.(\ref{eq:objBR2}) and \cite{jiang2017asymmetric}, $\mathbf{B}$ can be updated bit by bit. In other words, we update one column in $\mathbf{B}$ with remaining columns fixed. Let $\mathbf{B}_{*c}$ be the $c$-th column and $\mathbf{\hat{B}}_{c}$ be the remaining columns in $\mathbf{B}$. So do $\mathbf{U}_{*c}$, $\mathbf{\hat{U}}_{c}$, $\mathbf{V}_{*c}$, $\mathbf{\hat{V}}_{c}$, $\mathbf{Q}_{*c}$, and $\mathbf{\hat{Q}}_{c}$. Eq.(\ref{eq:objBR2}) can then be rewrote as:
\begin{equation}
\begin{aligned}
\min_{
\mathbf{B_{*c}}
}\textup{Tr}(\mathbf{B_{*c}}[2(\mathbf{U}_{*c}^{T}\mathbf{\hat{U}}_{c}+
\mathbf{V}_{*c}^{T}\mathbf{\hat{V}}_{c})\mathbf{\hat{B}}_{c}^{T}+\mathbf{Q}_{*c}^{T}]+\textup{const}\\ \ \textup{s.t.} \ \mathbf{B} \in \{-1,+1\}^{n\times k}
\end{aligned}\label{eq:objBR3}
\end{equation}
Obviously, the optimal solution for $\mathbf{B_{*c}}$ is
\begin{equation}
\begin{aligned}
\mathbf{B_{*c}} = -\textup{sign}(2\mathbf{\hat{B}}_{c}(\mathbf{\hat{U}}_{c}^{T}\mathbf{U}_{*c}+
\mathbf{\hat{V}}_{c}^{T}\mathbf{V}_{*c})+\mathbf{Q}_{*c})
\end{aligned}\label{eq:updateB}
\end{equation}
After computing $\mathbf{B_{*c}}$, we update $\mathbf{B}$ by replace the $c$-th column with $\mathbf{B_{*c}}$. Then we repeat Eq.(\ref{eq:updateB}) until all columns are updated.

Overall, the optimization of the proposed method is listed in Algorithm \ref{algorithm:dadhl}.

\begin{algorithm}[!tbp]
\caption{Dual Asymmetric Deep Hashing Learning (DADH)}
\begin{algorithmic}[1]
\renewcommand{\algorithmicrequire}{\textbf{Input:}}
\renewcommand{\algorithmicensure}{\textbf{End}}
\REQUIRE Training data $\mathbf{X}$/$\mathbf{Y}$; similarity matrix $\mathbf{S}$; hash code length $k$; predefined parameters $\tau$, $\gamma$ and $\eta$.
\renewcommand{\algorithmicrequire}{\textbf{Output:}}
\renewcommand{\algorithmicensure}{\textbf{End}}
\REQUIRE Hashing functions $\mathcal{F}$ and $\mathcal{G}$.
\renewcommand{\algorithmicrequire}{\textbf{Initialization:}}
\renewcommand{\algorithmicensure}{\textbf{End}}
\REQUIRE Initialize weights of the first seven layers by using the pretrained ImageNet model; the last layer is initialized randomly; $\mathbf{B}$ is set to be a matrix whose elements are zero.
\WHILE {not converged or not reach the maximum iteration}
\STATE
\textbf{Update $(\mathbf{F},\mathbf{W}_{f})$}:
\\Fix $(\mathbf{G},\mathbf{W}_{g})$ and $\mathbf{B}$ and update $(\mathbf{F},\mathbf{W}_{f})$ using back-propagation according to Eq.(\ref{eq:updateF}).
\STATE
 \textbf{Update $(\mathbf{G},\mathbf{W}_{g})$}:
 \\Fix $(\mathbf{F},\mathbf{W}_{f})$ and $\mathbf{B}$ and update $(\mathbf{G},\mathbf{W}_{g})$ using back-propagation according to Eq.(\ref{eq:updateG}).
 \STATE
  \textbf{Update $\mathbf{B}$}:
 \\Fix $(\mathbf{F},\mathbf{W}_{f})$ and $(\mathbf{G},\mathbf{W}_{g})$ and update $\mathbf{B}$ according to Eq.(\ref{eq:updateB}).
\ENDWHILE
\end{algorithmic}\label{algorithm:dadhl}
\end{algorithm}

\subsection{Query}
When $\mathbf{W}_{f}$ and $\mathbf{W}_{g}$ are learned, the hash functions corresponding to the two neural networks are subsequently obtained. For a given testing image $\mathbf{x}^{*}$, two kinds of binary codes can be computed, which are $\mathbf{b}_{f}^{*}=\textup{sign}(\mathbf{f}(\mathbf{x}^{*},\mathbf{W}_{f}))$ and $\mathbf{b}_{g}^{*}=\textup{sign}(\mathbf{f}(\mathbf{x}^{*},\mathbf{W}_{g}))$, respectively. Note that since $\textup{tanh}$ will not influence the sign of each element at the testing phase, we do not apply $\textup{tanh}$ for the output. From the experiments we find that the performances computed through the first and second networks are quite similar. To obtain a more robust result, we use the average of two outputs as the final result in our experiment.
\begin{equation}
\begin{aligned}
\mathbf{b}^{*} = \textup{sign}(0.5[\mathbf{f}(\mathbf{x}^{*},\mathbf{W}_{f})+\mathbf{g}(\mathbf{x}^{*},\mathbf{W}_{g})])
\end{aligned}\label{eq:query}
\end{equation}

\section{Experiments}
In this section, experiments are conducted on three large-scale datasets to demonstrate the effectiveness of the proposed method compared with some state-of-the-art approaches. We first describe the datasets used in our experiments, followed by the description of baselines, evaluation protocol, and implementation. We then make a comparison with other methods. The parameter sensitivity as well as the convergence are subsequently discussed.

\begin{table}\small
\caption{The MAP scores obtained by different methods on the IAPR TC-12 dataset.}
\centering
\begin{tabular}{c | c| c| c| c| c| c}
\hline
\multicolumn{1}{c|}{Method}&{8-bit}&{12-bit}&{16-bit}&{24-bit}&{36-bit}&{48-bit}\\
\hline
{LSH}&{32.89}&{33.39}&{33.25}&{34.33}&{34.99}&{35.15}\\
\hline
\hline
{ITQ}&{35.83}&{36.00}&{36.21}&{36.55}&{36.66}&{36.75}\\
\hline
{DPLM}&{36.38}&{36.86}&{37.35}&{37.71}&{38.64}&{38.83}\\
\hline
{SDH}&{36.92}  &{37.81}&{37.45}&{38.12}&{38.53}&{38.27}\\
\hline
{SGH}&{34.48}&{35.01}&{35.32}&{35.49}&{35.84}&{36.21}\\
\hline
\hline
{DPSH}&{45.19}&{46.03}&{46.82}&{47.37}&{47.97}&{48.60}\\
\hline
{ADSH}&{44.69}  &{46.98}&{48.25}&{49.06}&{50.24}&{50.59}\\
\hline
{DAPH}&{44.33}&{44.48}&{44.73}&{45.12}&{45.24}&{45.52}\\
\hline
\hline
{DADH}&\textbf{46.54}  &\textbf{49.27}&\textbf{50.83}&\textbf{52.71}&\textbf{54.47}&\textbf{55.39}\\
\hline
\end{tabular}\label{tab:IAPR}
\end{table}

\begin{table*}\small
\caption{The Top-500 MAP and Top-500 Precision scores obtained by different methods on the IAPR TC-12 dataset.}
\centering
\begin{tabular}{c | c| c| c| c| c| c||c|c|c|c|c|c}
\hline
\multicolumn{1}{c|}{Evaluation}&\multicolumn{6}{c||}{MAP@Top500}&\multicolumn{6}{c}{Precision@Top500}\\
\hline
\multicolumn{1}{c|}{Method}&{8-bit}&{12-bit}&{16-bit}&{24-bit}&{36-bit}&{48-bit}
&{8-bit}&{12-bit}&{16-bit}&{24-bit}&{36-bit}&{48-bit}\\
\hline
{LSH}&{36.19}&{37.47}&{37.53}&{39.83}&{40.50}&{41.52}    &{35.04}&{36.10}&{36.12}&{38.10}&{38.82}&{39.55}\\
\hline
\hline
{ITQ}&{41.16}&{41.91}&{42.76}&{43.67}&{44.07}&{44.30}    &{39.73}&{40.30}&{40.88}&{41.60}&{41.86}&{42.04}\\
\hline
{DPLM}&{42.21}&{43.12}&{43.89}&{44.50}&{45.86}&{46.42}    &{40.84}&{41.65}&{42.33}&{42.82}&{44.02}&{44.44}\\
\hline
{SDH}&{44.08}  &{45.31}&{46.00}&{47.30}&{48.17}&{48.06}    &{42.45}  &{43.54}&{43.94}&{45.00}&{45.79}&{45.49}\\
\hline
{SGH}&{40.04}&{41.54}&{42.09}&{42.48}&{43.07}&{43.97}    &{38.67}&{39.70}&{40.08}&{40.33}&{40.79}&{41.45}\\
\hline
\hline
{DPSH}&{57.07}&{58.13}&{59.94}&{61.61}&{63.05}&{64.49}    &{55.03}&{55.90}&{57.73}&{58.96}&{60.15}&{61.40}\\
\hline
{ADSH}&{53.70}  &{58.12}&{61.35}&{63.25}&{64.90}&{65.59}   &{52.31}&{56.47}&{58.97}&{60.29}&{61.94}&{62.50}\\
\hline
{DAPH}&{56.26}&{57.98}&{59.48}&{61.27}&{62.57}&{63.94}    &{54.32}&{55.44}&{56.52}&{57.92}&{58.95}&{60.03}\\
\hline
\hline
{DADH}&\textbf{57.12}  &\textbf{62.67}&\textbf{65.15}&\textbf{67.80}&\textbf{70.11}&\textbf{70.93}   &\textbf{55.31}&\textbf{60.20}&\textbf{62.52}&\textbf{64.93}&\textbf{67.13}&\textbf{67.81}\\
\hline
\end{tabular}\label{tab:IAPR500}
\end{table*}

\begin{figure*}
\begin{center}
   \includegraphics[width=0.95\linewidth]{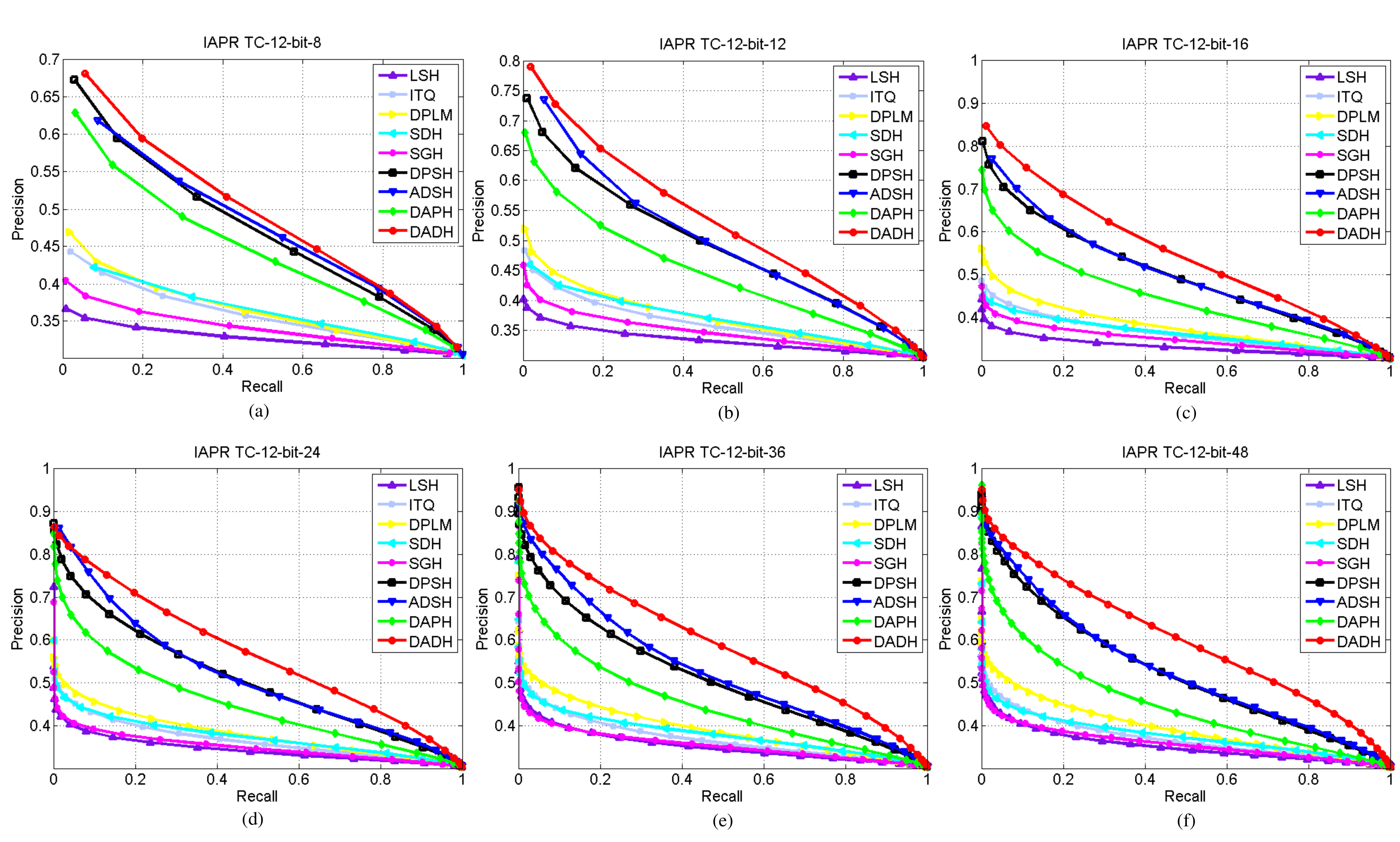}
\end{center}
   \caption{The Precision-Recall curves computed by LSH, ITQ, DPLM, SDH, SGH, DPSH, ADSH, DAPH, and DADH on the IAPR TC-12 dataset. Figures from (a) to (f) are associated with the code length 8-bit, 12-bit, 16-bit, 24-bit, 36-bit and 48-bit.}
\label{fig:PR_IAPR}
\end{figure*}

\subsection{Datasets}
Three datasets including IAPR TC-12 \cite{escalante2010segmented},   MIRFLICKR-25K \cite{huiskes2008mir} and CIFAR-10 \cite{krizhevsky2009learning} are used in this paper.

IAPR TC-12 \cite{escalante2010segmented} dataset consists of 20000 images associated with 255 categories. Since some samples have multiple labels, we set $S_{ij}=1$ only if there is at least one same label for the $i$-th and $j$-th sample. In this dataset, 2000 images are used for testing and 5000 samples selected from the remaining 18000 (retrieval set) points are used for training to greatly reduce the training time.

MIRFLICKR-25K dataset \cite{huiskes2008mir} is composed of 25000 images collected from the Flickr website. According to \cite{jiang2016deep}, 20015 images associated with 24 categories are selected. Being similar to IAPR TC-12 dataset, some images has multiple labels, we also define two images be a ground-truth neighbor if they share at least one same label. Additionally, 2000 images are randomly selected as the testing data, and the rest is defined as the retrieval data. Meanwhile, 5000 samples selected from the retrieval data are used for training.

CIFAR-10 dataset \cite{krizhevsky2009learning} contains 60000 32$\times$32 color images with ten categories. Each image belongs to one of these ten classes. Two images will be regarded as semantic neighbor if they have the same label. Being similar to the setting in \cite{jiang2017asymmetric}, we randomly select 1000 samples to be the testing data. In order to reduce the training time, we also randomly select 5000 images from the remaining 59000 images as the training data. The rest is then regarded as the retrieval set.

\subsection{Baseline and Evaluation Protocol}
To demonstrate the superiority of DADH, some existing hashing methods are used for comparison, including one data-independent methods (LSH \cite{gionis1999similarity}), four traditional data-dependent methods (ITQ \cite{gong2013iterative}, DPLM \cite{shen2016fast}, SDH \cite{shen2015supervised}, SGH \cite{jiang2015scalable}) and three deep learning based hashing methods (DPSH \cite{li2015feature}, ADSH \cite{jiang2017asymmetric}, DAPH \cite{shen2017deep}). Since LSH, ITQ, DPLM, SDH, and SGH are not deep learning methods, features should be extracted previously. For these three datasets, we have extracted the 4096-D CNN feature 512-D GIST feature, respectively. We have found that these five approaches often achieve a better performance on the GIST feature. Thus we use the GIST feature as the input for LSH, ITQ, DPLM, SDH, and SGH. For DPSH, ADSH, and DAPH, the raw image is used as the input and all images are resized into 224$\times$224$\times$3. For all deep learning methods, the CNN-F is used as the network for feature extraction and the parameters in DPSH and ADSH are set according to their descriptions in their publications. Note that, since the code for DAPH is not released, we implement it with the deep learning toolbox MatConvNet \cite{vedaldi2015matconvnet} very carefully. Additionally, the original network structure in DAPH is not CNN-F which means the parameters in \cite{shen2017deep} may be not optimal. Thus, we try our best to tune the parameters in DAPH.

To quantatively measure the proposed method and other comparison methods, two widely used metrics containing mean average precision (MAP) and precision-recall (PR) are adopted. The definitions of MAP criteria is demonstrated as follows: Given a query, the average precision (AP) is first computed by searching a set of $R$ retrieved results.
\begin{equation}
\begin{aligned}
AP = \frac{1}{T}\sum_{r=1}^{R}P(r)\delta(r)
\end{aligned}
\end{equation}
where $T$ is the total number of document set in retrieved set, $P(r)$ is the precision of top $r$ retrieved cases, and $\delta(r)$ denotes whether the retrieved sample is relevant (if the instance is a true neighbor of the query, $\delta(r)=1$, otherwise $\delta(r)=0$). Additionally, being similar to some existing methods \cite{shen2017deep} \cite{jiang2016deep}, Top-500 MAP and Top-500 Precision are also exploited to evaluate the superiority of the proposed method.

\subsection{Implementation}
We implement DADH with the deep learning toolbox MatConvNet \cite{vedaldi2015matconvnet} on Titan X GPU. The pre-trained ImageNet model is used to initialize the first seven layers in each stream and the weights in the last layer are initialized randomly. During the training time, we set the mini-batch size to be 64 and divide the learning rate among $[10^{-6},10^{-4}]$ into 150 iterations. In other words, the learning rate gradually reduces from $10^{-4}$ to $10^{-6}$ and the stochastic gradient descent is used to update the weights. Based on the cross-validation (a small set for validation is randomly selected from the training data), we set $\gamma=100$, $\eta=10$, and $\tau=10$ in the three datasets. We will further demonstrate the insensitivity of these parameters in the following subsection.

\subsection{Comparison with Other Methods}
\subsubsection{IAPR TC-12}

\begin{table}\small
\caption{The MAP scores obtained by different methods on the MIRFLICKR-25K dataset.}
\centering
\begin{tabular}{c | c| c| c| c| c| c}
\hline
\multicolumn{1}{c|}{Method}&{8-bit}&{12-bit}&{16-bit}&{24-bit}&{36-bit}&{48-bit}\\
\hline
{LSH}&{56.06}&{56.10}&{56.72}&{56.82}&{57.56}&{57.35}\\
\hline
\hline
{ITQ}&{57.59}&{57.57}&{57.70}&{57.79}&{57.84}&{57.87}\\
\hline
{DPLM}&{60.42}&{60.51}&{60.53}&{60.70}&{60.91}&{60.79}\\
\hline
{SDH}&{60.17}  &{60.27}&{60.46}&{60.67}&{60.96}&{61.59}\\
\hline
{SGH}&{57.35}  &{57.54}&{57.57}&{57.67}&{57.80}&{57.86}\\
\hline
\hline
{DPSH}&{73.48}  &{74.68}&{75.58}&{76.01}&{76.09}&{76.05}\\
\hline
{ADSH}&{75.39}  &{76.41}&{76.98}&{76.59}&{76.20}&{74.53}\\
\hline
{DAPH}&{72.79}  &{74.70}&{74.30}&{74.14}&{73.81}&{73.41}\\
\hline
\hline
{DADHL}&\textbf{77.15}  &\textbf{78.16}&\textbf{78.64}&\textbf{79.44}&\textbf{79.72}&\textbf{79.26}\\
\hline
\end{tabular}\label{tab:FLICK}
\end{table}

\begin{table*}\small
\caption{The Top-500 MAP and Top-500 Precision scores obtained by different methods on the MIRFLICKR-25K dataset.}
\centering
\begin{tabular}{c | c| c| c| c| c| c||c|c|c|c|c|c}
\hline
\multicolumn{1}{c|}{Evaluation}&\multicolumn{6}{c||}{MAP@Top500}&\multicolumn{6}{c}{Precision@Top500}\\
\hline
\multicolumn{1}{c|}{Method}&{8-bit}&{12-bit}&{16-bit}&{24-bit}&{36-bit}&{48-bit}
&{8-bit}&{12-bit}&{16-bit}&{24-bit}&{36-bit}&{48-bit}\\
\hline
{LSH}&{57.57}&{58.09}&{59.27}&{59.55}&{60.84}&{60.67}    &{56.94}&{57.25}&{58.50}&{58.69}&{59.93}&{59.84}\\
\hline
\hline
{ITQ}&{61.00}&{61.17}&{61.32}&{61.61}&{61.64}&{61.85}    &{59.98}&{60.11}&{60.29}&{60.55}&{60.56}&{60.74}\\
\hline
{DPLM}&{63.08}&{63.84}&{64.05}&{64.36}&{64.73}&{65.03}    &{62.36}&{63.14}&{63.31}&{63.64}&{64.05}&{64.28}\\
\hline
{SDH}&{65.16}  &{64.75}&{65.40}&{65.34}&{65.57}&{66.46}    &{64.05}  &{63.93}&{64.58}&{64.44}&{64.80}&{65.60}\\
\hline
{SGH}&{60.74}  &{61.23}&{61.48}&{61.42}&{61.91}&{62.04}    &{59.81}  &{60.21 }&{60.46}&{60.42}&{60.85}&{60.93}\\
\hline
\hline
{DPSH}&{82.88}&{83.84}&{84.34}&{84.84}&{85.77}&{85.64}    &{81.85}&{83.01}&{83.58}&{84.11}&{84.97}&{84.80}\\
\hline
{ADSH}&{82.14}&{83.80}&{84.94}&{84.90}&{84.20}&{82.06}    &{81.50}&{83.18}&{84.15}&{84.03}&{83.52}&{81.31}\\
\hline
{DAPH}&{81.08}&{84.20}&{83.71}&{84.45}&{84.02}&{84.07}    &{80.37}&{83.24}&{82.73}&{83.40}&{82.93}&{82.91}\\
\hline
\hline
{DADHL}&\textbf{85.80}&\textbf{86.83}&\textbf{86.90}&\textbf{87.42}&\textbf{87.98}&\textbf{87.58}    &\textbf{84.73}&\textbf{85.78}&\textbf{85.76}&\textbf{86.68}&\textbf{87.08}&\textbf{86.80}\\
\hline
\end{tabular}\label{tab:FLICK500}
\end{table*}

\begin{figure*}
\begin{center}
   \includegraphics[width=0.95\linewidth]{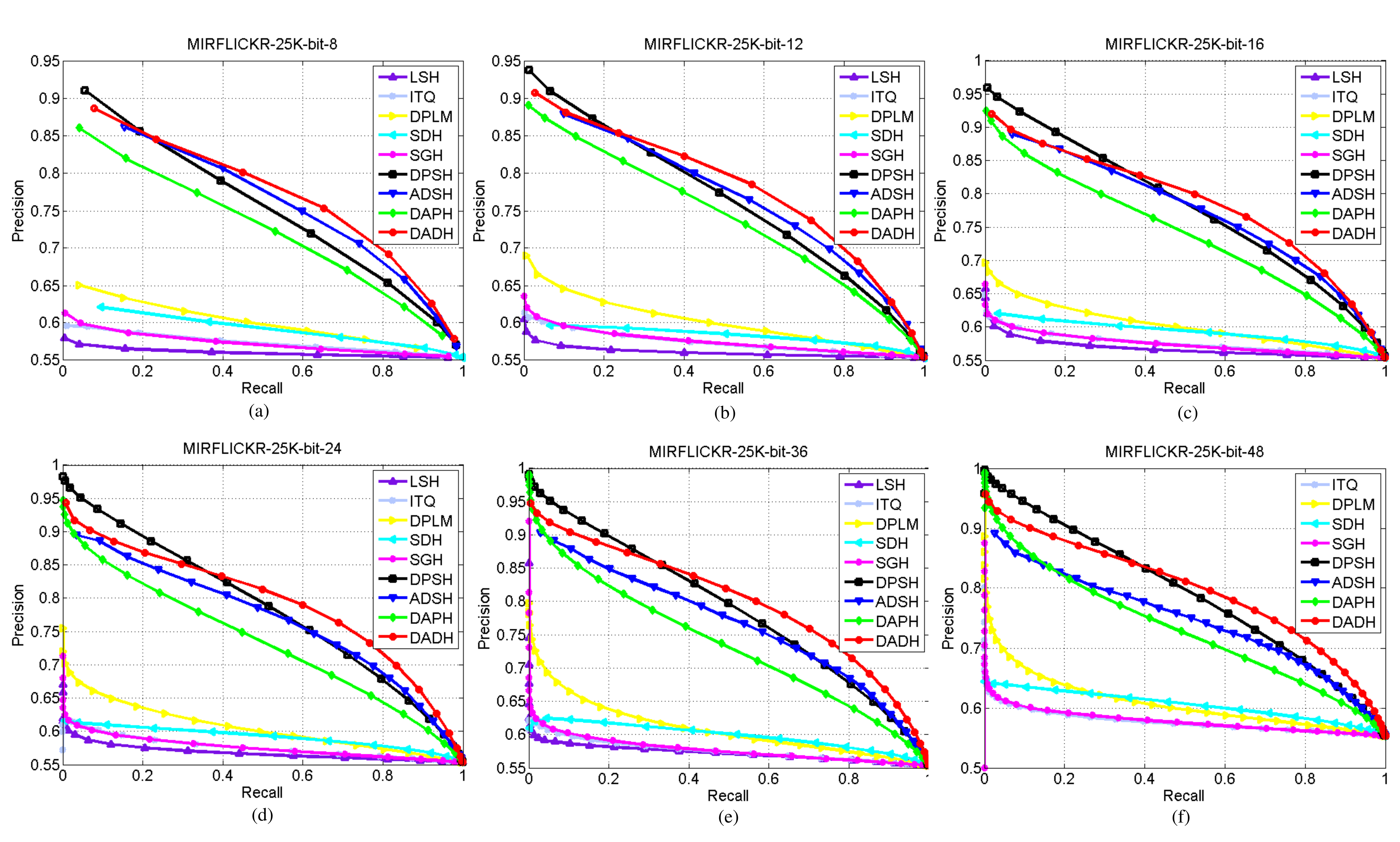}
\end{center}
   \caption{The Precision-Recall curves computed by LSH, ITQ, DPLM, SDH, SGH, DPSH, ADSH, DAPH, and DADH on the MIRFLICKR-25K dataset. Figures from (a) to (f) are associated with the code length 8-bit, 12-bit, 16-bit, 24-bit, 36-bit and 48-bit.}
\label{fig:PR_FLICK}
\end{figure*}

The MAP scores obtained by different methods on the IAPR TC-12 dataset are shown in Tab.\ref{tab:IAPR}. It is easy to observe that DADH achieves a remarkable improvement in MAP scores compared with other approaches. In contrast to the data-independent method LSH, DADH achieves more than 10\%-20\% percents higher in MAP scores. Compared with ITQ, DPLM, SDH and SGH, there is also an obvious enhancement. Specifically, our proposed method obtains at least 46.54\% MAP score and reaches as high as 55.39\% when the bit length is 48, while the best result obtained by ITQ, DPLM, SDH and SGH is only 38.83\%, being far below than our's. Referring to DPSH, ADSH and DAPH, DADH also has more or less improvement in MAP scores. Particularly, compared with DAPH, the presented approach gains about or more than 5\% enhancement when the bit length ranges from 12 to 48. In comparison to DPSH and ADSH, the performance obtained by DADH also has about 3\%-5\% improvement when the length of hashing bit is 24, 36, and 48, respectively.

The Top-500 MAP and Top-500 Precision scores on the IAPR TC-12 dataset are listed in Tab.\ref{tab:IAPR500}. From this table we can see that the experimental results obtained by the deep learning based methods including DPSH, ADSH, DAPH and DADH are remarkably better than that computed by other traditional approaches. Specifically, there is about 10\%-20\% improvement in MAP@Top500 and Precision@Top500 scores in most cases. In contrast to DPSH, ADSH and DAPH, the performance achieved by the proposed method DADH still reaches the best point. Except the case when the bit length is 8, DADH always gain 4\% or more enhancement in MAP@Top500 and Precision@Top500 scores, indicating the effectiveness of our method.

The Precision-Recall curves computed by different methods on the IAPR TC-12 dataset are displayed in Fig.\ref{fig:PR_IAPR}, when the bit length changes from 8 to 48. We can easily observe that covered areas gained by DADH are much larger than that obtained by other comparison methods. We can find that the proposed method can dramatically outperform the traditional data-independent and data-dependent strategies. Referring to DPSH, ADSH and DAPH, there is also a better achievement in all cases with different values of the code length.

\subsubsection{MIRFLICKR-25K}

The MAP results of the experiment conducted on the MIRFLICKR-25K dataset are tabulated in Tab.\ref{tab:FLICK}. We can see that DADH achieves the best performance in all cases with different values of the code length. Being similar to the results on IAPR TC-12 dataset, DPLM, SDH, DPSH, ADSH, DAPH and DADH can obtain higher values in MAP compared with LSH, ITQ and SGH. Referring to the comparison between the traditional methods and deep learning methods, DPSH, ADSH, DAPH and DADH dramatically outperform LSH, ITQ, DPLM, SDH and SGH. Make a comparison between the proposed method with other deep hashing approaches, DADH still has a more or less improvement. For DADH, there is about 1.5\%-3\% enhancement on MAP scores compared with these three deep hashing approaches.

\begin{table}\small
\caption{The MAP scores obtained by different methods on the CIFAR-10 dataset.}
\centering
\begin{tabular}{c | c| c| c| c| c| c}
\hline
\multicolumn{1}{c|}{Method}&{8-bit}&{12-bit}&{16-bit}&{24-bit}&{36-bit}&{48-bit}\\
\hline
{LSH}&{14.19}&{13.26}&{13.13}&{13.84}&{14.90}&{15.13}\\
\hline
\hline
{ITQ}&{16.57}&{17.12}&{16.94}&{17.08}&{17.38}&{17.58}\\
\hline
{DPLM}&{21.97}&{22.76}&{23.91}&{25.89}&{27.53}&{28.85}\\
\hline
{SDH}&{30.93}  &{32.50}&{33.59}&{35.36}&{35.59}&{36.53}\\
\hline
{SGH}&{15.06}  &{15.43}&{15.64}&{16.07}&{16.78}&{16.88}\\
\hline
\hline
{DPSH}&{63.48}  &{66.97}&{68.83}&{73.45}&{74.66}&{75.02}\\
\hline
{ADSH}&{56.67}  &{71.41}&{76.50}&{80.40}&{82.73}&{82.73}\\
\hline
{DAPH}&{59.09}  &{61.17}&{68.15}&{69.22}&{70.74}&{70.28}\\
\hline
\hline
{DADHL}&\textbf{71.86}  &\textbf{75.12}&\textbf{80.33}&\textbf{81.70}&\textbf{83.16}&\textbf{83.90}\\
\hline
\end{tabular}\label{tab:CIFAR}
\end{table}

\begin{table*}\small
\caption{The Top-500 MAP and Top-500 Precision scores obtained by different methods on the CIFAR-10 dataset.}
\centering
\begin{tabular}{c | c| c| c| c| c| c||c|c|c|c|c|c}
\hline
\multicolumn{1}{c|}{Evaluation}&\multicolumn{6}{c||}{MAP@Top500}&\multicolumn{6}{c}{Precision@Top500}\\
\hline
\multicolumn{1}{c|}{Method}&{8-bit}&{12-bit}&{16-bit}&{24-bit}&{36-bit}&{48-bit}
&{8-bit}&{12-bit}&{16-bit}&{24-bit}&{36-bit}&{48-bit}\\
\hline
{LSH}&{20.29}&{19.92}&{19.81}&{21.17}&{23.07}&{24.42}    &{14.54}&{15.26}&{16.19}&{18.13}&{20.45}&{21.44}\\
\hline
\hline
{ITQ}&{24.56}&{27.72}&{27.55}&{28.65}&{29.54}&{30.24}    &{18.63}&{21.88}&{22.53}&{24.59}&{25.65}&{26.40}\\
\hline
{DPLM}&{29.02}&{34.07}&{35.66}&{38.86}&{40.28}&{41.67}    &{24.61}&{29.72}&{31.93}&{35.62}&{37.98}&{39.84}\\
\hline
{SDH}&{28.34}  &{34.78}&{37.92}&{41.69}&{42.77}&{44.57}    &{27.91}  &{33.83}&{37.31}&{42.39}&{43.78}&{45.08}\\
\hline
{SGH}&{25.92}  &{25.11}&{25.91}&{27.13}&{28.76}&{29.46}    &{19.06}  &{20.10}&{21.53}&{23.46}&{24.85}&{25.42}\\
\hline
\hline
{DPSH}&{58.58}  &{66.85}&{71.48}&{75.74}&{79.69}&{80.62}    &{64.13}  &{70.80}&{74.71}&{78.34}&{80.59 }&{81.55}\\
\hline
{ADSH}&{58.92}  &{70.07}&{74.95}&{78.09}&{78.85}&{77.61}    &{61.09}  &{73.71}&{78.85}&{81.84}&{83.45}&{82.86}\\
\hline
{DAPH}&{50.88}  &{66.24}&{72.43}&{77.31}&{79.21}&{80.40}    &{54.22}  &{68.28}&{74.42}&{77.36}&{78.80}&{79.55}\\
\hline
\hline
{DADHL}&\textbf{67.11}  &\textbf{73.08}&\textbf{78.42}&\textbf{82.02}&\textbf{83.51}&\textbf{84.17}    &\textbf{72.95}  &\textbf{77.53}&\textbf{82.68}&\textbf{84.18 }&\textbf{85.23}&\textbf{85.59}\\
\hline
\end{tabular}\label{tab:CIFAR500}
\end{table*}
\begin{figure*}
\begin{center}
   \includegraphics[width=0.95\linewidth]{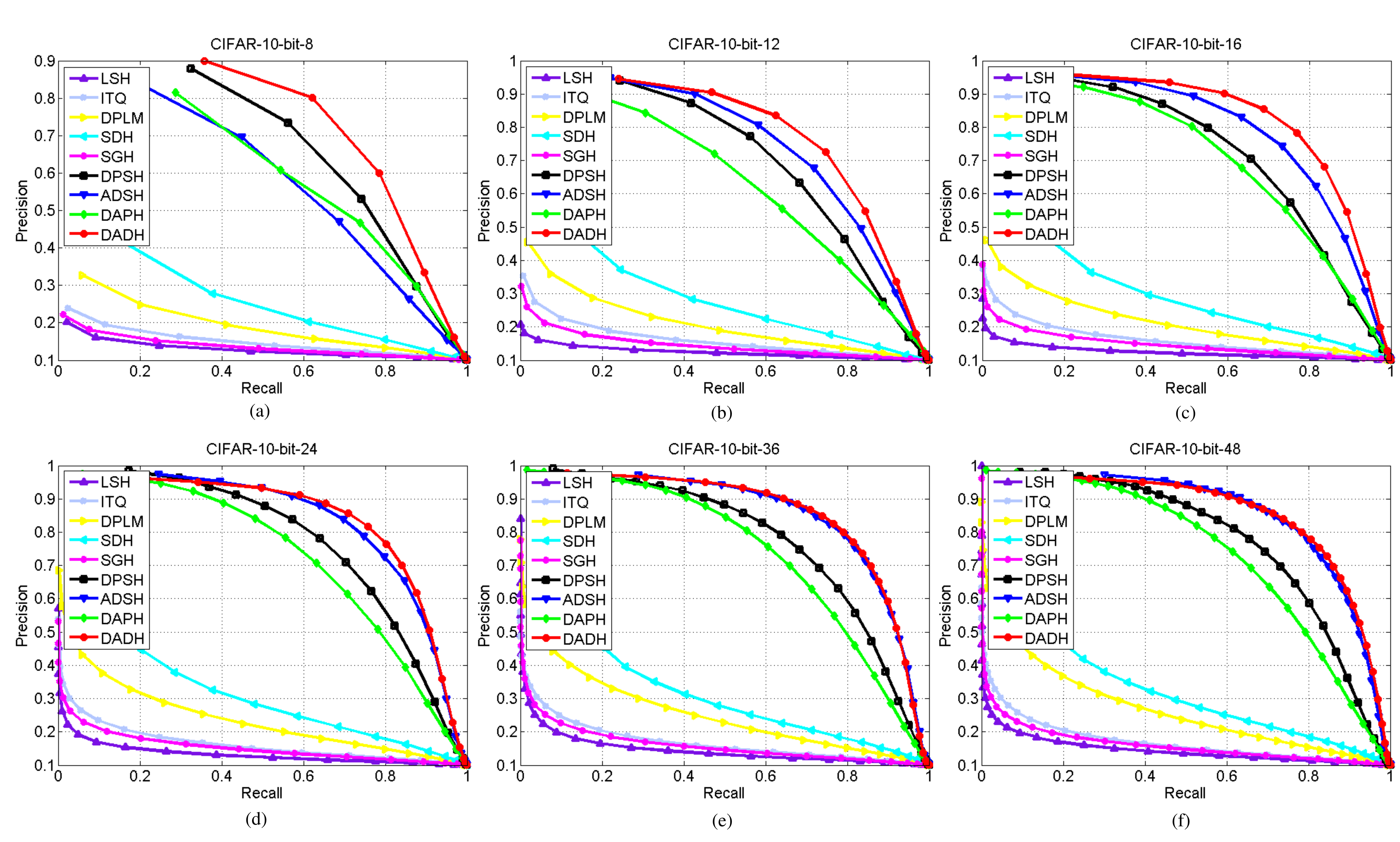}
\end{center}
   \caption{The Precision-Recall curves computed by LSH, ITQ, DPLM, SDH, SGH, DPSH, ADSH, DAPH, and DADH on the CIFAR-10 dataset. Figures from (a) to (f) are associated with the code length 8-bit, 12-bit, 16-bit, 24-bit, 36-bit and 48-bit.}
\label{fig:PR_CIFAR}
\end{figure*}

The Top-500 MAP and Top-500 Precision scores on the MIRFLICKR-25K dataset are displayed in Tab.\ref{tab:FLICK500}. It is easy to observe that DADH obtains the best performance in both MAP@Top500 and Precision@Top500, demonstrating the superiority compared with other existing strategies. With the change of the code length, the MAP@Top500 and Precision@Top500 increase from (85.80\%, 84.73\%) to (87.58\%, 86.80\%), while the highest values obtained by LSH, ITQ and SGH are only (62.04\%, 60.93\%), being much lower than ours'. In contrast to DPLM and SDH, our strategy still gains more than 20\% enhancement in most cases. Furthermore, results computed by DADH are much higher than that calculated by DPSH, ADSH and DAPH. Concretely, scores of Top-500 MAP and Top-500 Precision gained by DADH are almost always higher than 85\%, while these scores calculated by other deep hashing methods are below than 85\% in most cases.

\begin{figure*}
\begin{center}
   \includegraphics[width=0.95\linewidth]{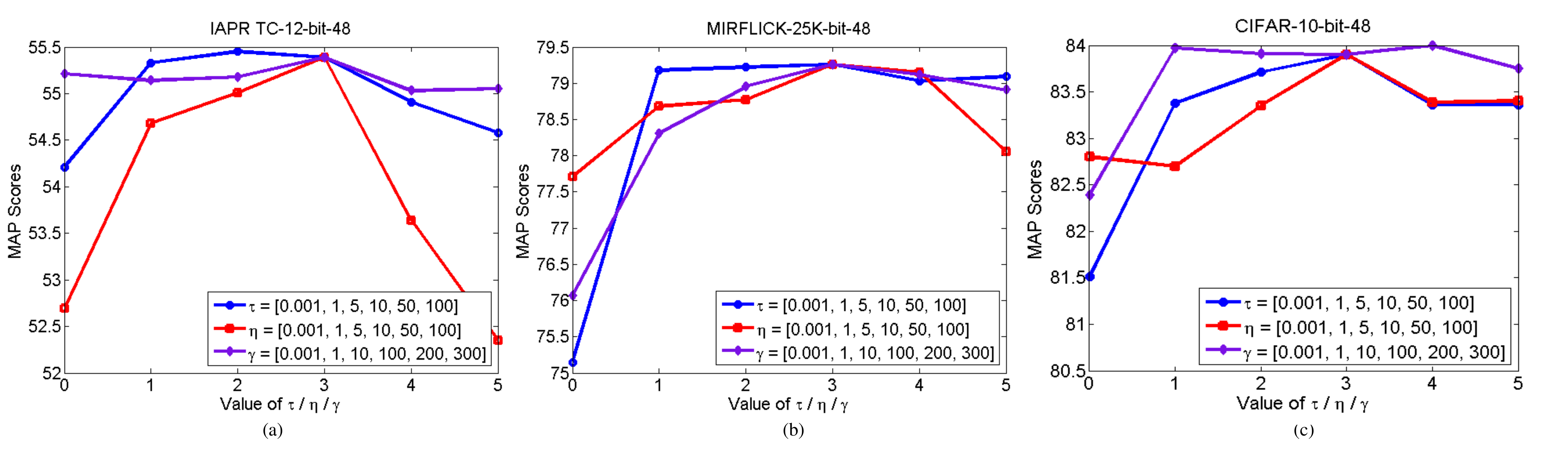}
\end{center}
   \caption{The MAP scores with the change of parameter $\tau$, $\eta$ and $\gamma$ on three datasets, when the code length is 48.}
\label{fig:param}
\end{figure*}

Fig.\ref{fig:PR_FLICK} show the Precision-Recall curves computed by different methods on the MIRFLICKR-25K dataset, when the bit length changes from 8 to 48. Note that we do not depict the Precision-Recall curve obtained by LSH in Fig.\ref{fig:PR_FLICK}(f), since its precision scores is far below than that of others. From Fig.\ref{fig:PR_FLICK} we can observe that DADH remarkably outperforms LSH, ITQ, DPLM, SDH, SGH, ADSH, and DAPH. Referring the comparison between DPSH and DADH, the proposed method is obviously superior to DPSH when the code length is 8 and 12, respectively. Although DPSH covers more areas when the recall value is smaller than 0.4 in Fig.\ref{fig:PR_FLICK}(c)-(f), it is inferior to DADH with the increase of the recall value. Overall, our method still outperforms DPSH when the code length is 16, 24, 36 and 48.

\begin{figure*}
\begin{center}
   \includegraphics[width=0.95\linewidth]{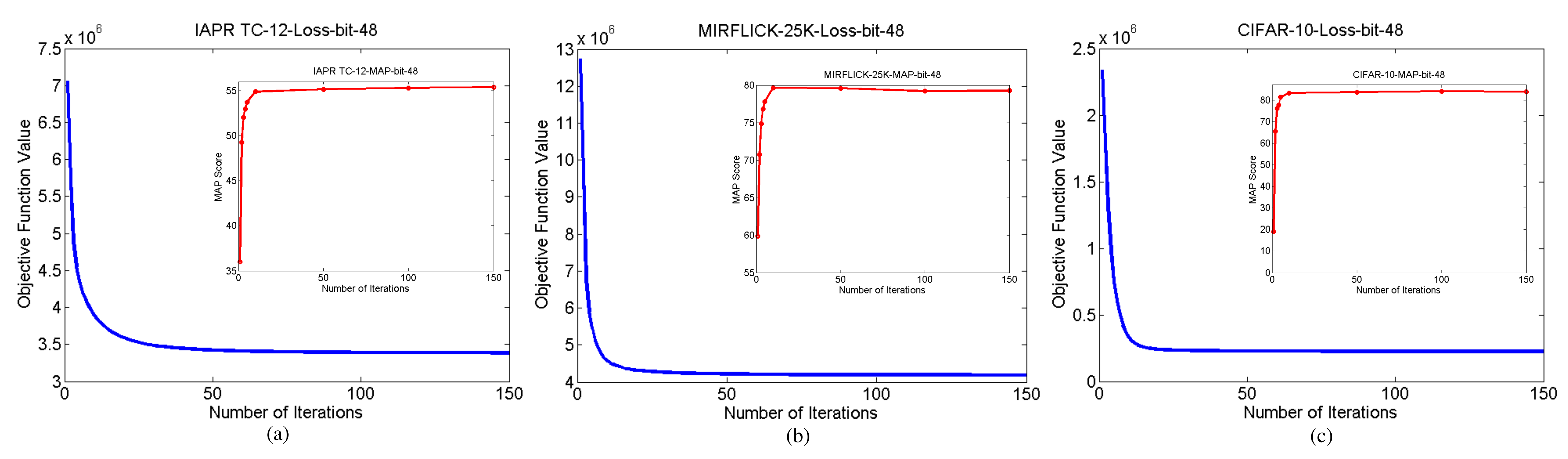}
\end{center}
   \caption{The change of objective function values and MAP scores with the increase of iterations.}
\label{fig:convergence}
\end{figure*}

\begin{figure*}
\begin{center}
   \includegraphics[width=0.95\linewidth]{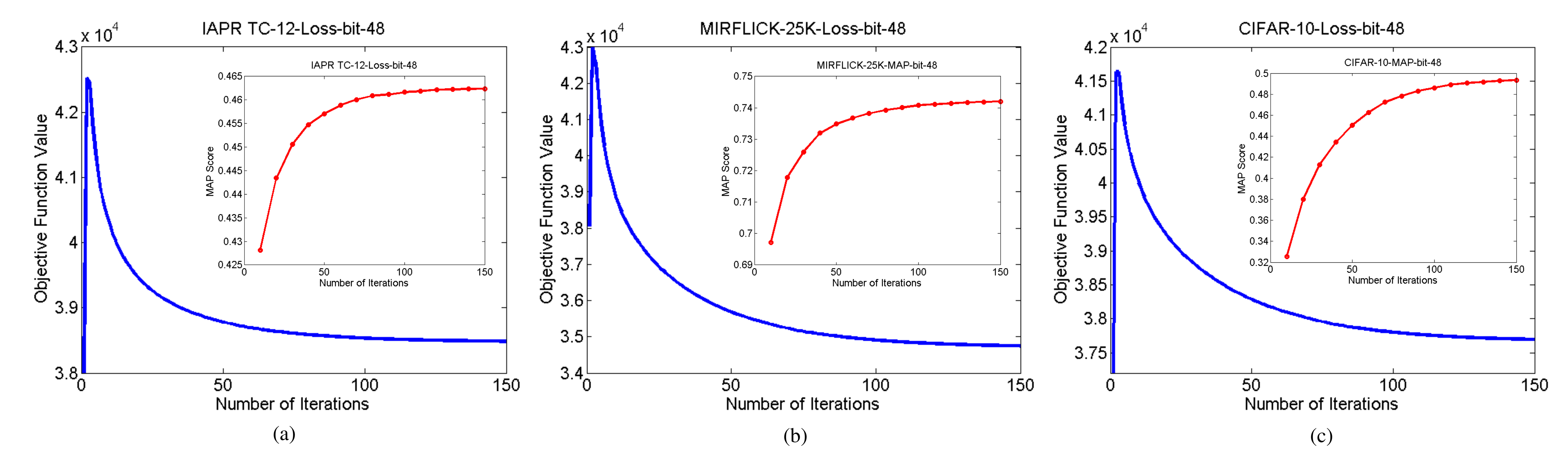}
\end{center}
   \caption{The change of objective function values and MAP scores with the increase of iterations. Note that, the asymmetric terms $\left \| \textup{tanh}(\mathbf{F})\mathbf{B}^{T} - k\mathbf{S} \right \|_{F}^{2}$ and $\left \| \textup{tanh}(\mathbf{G})\mathbf{B}^{T} - k\mathbf{S} \right \|_{F}^{2}$ are removed from the objective function.}
\label{fig:convcomp}
\end{figure*}

\subsubsection{CIFAR-10}
Tab.\ref{tab:CIFAR} lists the MAP scores obtained by the proposed method and various comparison approaches on the CIFAR-10 dataset. With the change of the code length from 8 to 48, the MAP scores computed by DADH rise from 71.86\% to 83.90\%, being much higher than that obtained by traditional hashing approaches, including LSH, ITQ, DPLM, SDH and SGH. In contrast to DPSH and DAPH, it is easy to observe that the presented method can achieve a better performance under the different code length. Except the cases when code length is 8 and 12, the MAP scores gained by DADH are always higher than 80\%, while the best performance computed by DPSH and DAPH is only 75.02\%. Also, ADSH is inferior to the proposed method, especially when the code length is small. This relatively indicates the effectiveness of our method no matter the code length is small or large.

The Top-500 MAP and Top-500 Precision scores computed by different methods on the CIFAR-10 dataset are shown in Tab.\ref{tab:CIFAR500} under the various code length. Obviously, four deep hashing methods always achieve dramatic experimental results compared with rest traditional approaches. The comparison between DADH and other deep hashing methods also substantiates the superiority of the proposed strategy. In contrast to DPSH, ADSH and DAPH, DADH has about 3\%-4\% enhancement in both Top-500 MAP and Top-500 Precision scores when the code length ranges from 12 to 48.

The Precision-Recall curves computed by different methods on the CIFAR-10 dataset are depicted in Fig.\ref{fig:PR_CIFAR} under the bit length ranging from 8 to 48. When the code length is 8, 12, 16, and 24, the Precision-Recall curves obtained by DADH covers the most areas. Although the performance computed by ADSH is competitive to ours' when the code length is 36 and 48, DADH is still much superior to LSH, ITQ, DPLM, SDH, SGH, DPSH, and DAPH.

\subsection{Parameter Sensitivity Analysis}

The MAP scores under the changes of different values of $\tau$, $\eta$ and $\gamma$ are shown in Fig.\ref{fig:param}. Note, we tune a parameter with others fixed. For instance, we tune $\tau$ in the range of $[0.001, 1, 5, 10, 50, 100]$ by fixing $\eta=10$ and $\gamma=100$, respectively. Similarly, we set $\tau=10$, $\gamma=100$ in $\eta$ tuning and $\tau=10$, $\eta=10$ in $\gamma$ tuning. As we can see, our model is insensitive to parameters. Specifically, $\tau$, $\eta$ and $\gamma$ have a wide range [1,50], [1,50] and [1,300], respectively. Our method always achieves a satisfactory performance when $\tau$, $\eta$ and $\gamma$ are in these ranges. This relatively demonstrates the robustness and effectiveness of the proposed method.
\subsection{Convergence Analysis}

To be honest, our proposed model can get a convergence with a few of iterations. The change of the objective function values and MAP scores on three datasets are displayed in Fig.\ref{fig:convergence} when the code length is 48-bit. It is easy to observe that DADH converges to a stable value after less than 30 iterations. In fact, we further find that the asymmetric terms $\left \| \textup{tanh}(\mathbf{F})\mathbf{B}^{T} - k\mathbf{S} \right \|_{F}^{2}$ and $\left \| \textup{tanh}(\mathbf{G})\mathbf{B}^{T} - k\mathbf{S} \right \|_{F}^{2}$ greatly contribute to the quick convergence. We try to remove these two terms from our objective function to study the influence. Note that, if the asymmetric terms are removed, the binary code $\mathbf{B}$ are updated through $\textup{sign}(\gamma[\textup{tanh}(\mathbf{G})+\textup{tanh}(\mathbf{F})])$. As shown in Fig.\ref{fig:convcomp}, if we remove $\left \| \textup{tanh}(\mathbf{F})\mathbf{B}^{T} - k\mathbf{S} \right \|_{F}^{2}$ and $\left \| \textup{tanh}(\mathbf{G})\mathbf{B}^{T} - k\mathbf{S} \right \|_{F}^{2}$ from the objective function, not only the MAP scores meet a degradation, but also our model converges much slower compared with the original DADH, indicating the necessity and significance of the asymmetric terms.

\section{Conclusion}
In this paper, we propose a novel deep hashing method named dual asymmetric deep hashing learning (DADH) for image retrieval. Specifically, two asymmetric networks are designed to integrate the feature representation and hash function learning into the end-to-end framework. A pairwise loss is introduced to exploit the semantic structure between each pair outputs. Furthermore, another pairwise loss is proposed to not only capture the similarity between the discrete binary codes and learned real-value features, but also contribute to a quick convergence at the training phase. Experiments are conducted on three large-scale datasets and the outstanding results substantiate the superiority of the proposed method.

\section*{Acknowledgment}
The work is partially supported by the GRF fund from the HKSAR Government, the central fund from Hong Kong Polytechnic University, the NSFC fund (61332011, 61272292, 61271344, 61602540), Shenzhen Fundamental Research fund (JCYJ20150403161923528, JCYJ20140508160910917), and the Science and Technology Development Fund (FDCT) of Macau 124/2014/A3.

\end{document}